\documentclass[10pt,twocolumn,letterpaper]{article}

\usepackage[pagenumbers]{cvpr} 

\usepackage{graphicx}
\usepackage{amsmath}
\usepackage{amssymb}
\usepackage{booktabs}
\usepackage{textcomp}
\usepackage{xcolor}
\usepackage{cite}
\usepackage{indentfirst}
\usepackage{csquotes}
\usepackage{multirow}
\usepackage{tabularx}
\usepackage{float}
\usepackage{color}
\usepackage{threeparttable}
\usepackage{soul}
\usepackage{pifont}
\usepackage{algorithm}
\usepackage{algpseudocode}
\usepackage{bigstrut}
\usepackage{makecell}
\usepackage{setspace}
\usepackage{bbding}

\definecolor{cvprblue}{rgb}{0.21,0.49,0.74}
\usepackage[pagebackref,breaklinks,colorlinks,citecolor=cvprblue]{hyperref}

\def\paperID{3844} 
\def\confName{CVPR}
\def\confYear{2024}

\title{IRConStyle: Image Restoration Framework Using Contrastive Learning and Style Transfer}

\author{Dongqi Fan, Xin Zhao, Liang Chang\\
University of Electronic Science and Technology of China\\
}

\begin{document}
\maketitle
\begin{abstract}
Recently, the contrastive learning paradigm has achieved remarkable success in high-level tasks such as classification, detection, and segmentation. However, in low-level tasks, like image restoration, contrastive learning is limited, and its effectiveness is uncertain. This raises a question: Why does the contrastive learning paradigm not yield remarkable results in image restoration? In this paper, we conduct in-depth analyses and propose three guidelines to address the above question. In addition, inspired by style transfer and based on contrastive learning, we propose a novel module for image restoration called \textbf{ConStyle}, which can be efficiently integrated into any U-Net structure network. By leveraging the flexibility of ConStyle, we develop a \textbf{general restoration network} for image restoration. The ConStyle and general restoration network together form an image restoration framework, namely \textbf{IRConStyle}. To demonstrate the capability and compatibility of ConStyle, we replace the general restoration network with transformer-based, CNN-based, and MLP-based networks, respectively. We perform extensive experiments on various image restoration tasks, including denoising, deblurring, deraining, and dehazing. The results on 19 benchmarks demonstrate that ConStyle can be integrated with any U-Net-based network and significantly enhance the performance of the network. For instance, ConStyle NAFNet significantly outperforms the original NAFNet on SOTS outdoor (dehazing) and Rain100H (deraining) datasets, with PSNR improvements of 4.16 dB and 3.58 dB with 85\% fewer parameters. The code is available at: \href{https://github.com/Dongqi-Fan/IRConStyle}{https://github.com/Dongqi-Fan/IRConStyle}
\end{abstract}

\section{Introduction}
\label{sec:intro}

\begin{figure}[htbp]
\centering
\includegraphics[width=\linewidth]{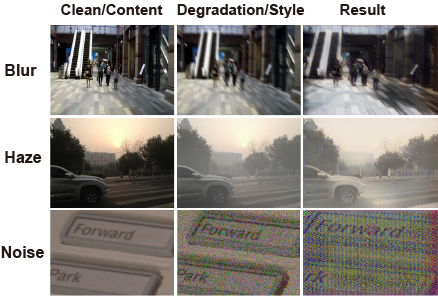}
\caption{The style transfer experiment by StyleFlow \cite{19fan2022styleflow} is performed to observe whether specific degradation of the degraded image could be transferred to a clean image.}     
\label{fig:fig1}
\end{figure}

Image Restoration (IR) aims to reconstruct degraded images into clean images, which can be applied in various fields such as surveillance, medical images, and autonomous driving. Recently, various IR works \cite{38liang2021swinir,29xie2021finding,40li2020lapar,35zhang2022practical,24valanarasu2022transweather,20li2020all,25park2023all,33poirier2023robust,37huang2023contrastive} have shown brilliant results and great prospects. However, these methods only focus on the specific module and operator design, ignoring the importance of overall network architecture and training strategy.

Typically, contrastive Learning (CL) paradigm\footnote{In this paper, the CL paradigm refers to the use of contrastive loss, additional data structures to store positive or negative samples, and a large number of samples in calculating contrastive loss.} \cite{1momentum,2improved,3bootstrap,4Chen_2021_ICCV, 5caron2021emerging,6chen2020big,11chen2020simple} is considered a type of self-supervised or unsupervised learning, which has shown great potential in high-level tasks, often achieving or even surpassing the performance of supervised counterpart in downstream tasks. However, to the best of our knowledge, only two works \cite{12wang2021unsupervised,13li2022all} have adopted the CL paradigm in the IR domain. Most studies \cite{14wang2021towards,32du2020learning,30chen2022learning,18moon2022feature,28chen2022unpaired,34chen2022unpaired} only use the contrastive loss. Note that whether these works adopt the CL paradigm or not, the final effect is not expected. \textbf{This raises a question: Why does the CL paradigm not achieve great success in IR?} For this, we conduct in-depth thinking and propose three guidelines to fully leverage the potential of the CL paradigm in IR (see \cref{Analysis} for details): Firstly, additional data structures should be used to store enough positive/negative samples. Secondly, it is necessary to fully use the encoder's intermediate feature map and the output latent code (the intermediate feature map and latent code referred to as \textbf{latent feature}). Finally, a reasonable pretext task should be adopted.

\begin{figure*}[htbp]
\centering
\includegraphics[width=\linewidth]{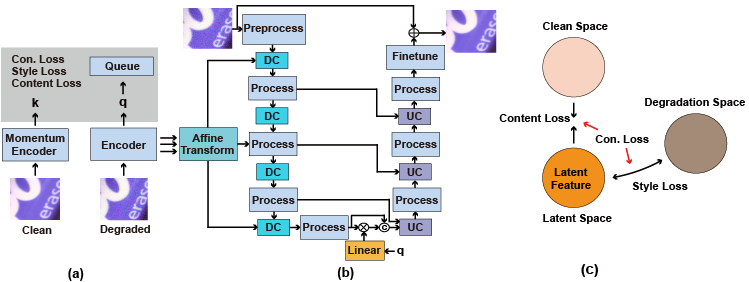}
\caption{The overall architecture of IRConStyle ((a)(b)). (a) ConStyle. (b) General restoration network, where the Preprocess, Process, and Finetune modules can be replaced with any specific operator. (c) Under the influence of Con. (contrastive), content, and style loss, the latent feature keeps moving closer to clean space and away from degradation.}
\label{fig:fig2}
\end{figure*}

Image style transfer refers to transferring the style of a style image to a content image so that the content image retains its original content while also adopting the style of the style image. Since style transfer can transfer a specific image style, \textbf{we wondered if we could treat degradations, such as blur and haze, as a kind of style and prevent the style image from transferring these degradations to the content image during IR.} To test our idea, based on StyleFlow \cite{19fan2022styleflow}, we conduct a transfer experiment (\cref{fig:fig1}) using the clean image as the content image and the degraded image as the style image. The noise, blur, and haze are treated as separate styles. \cref{fig:fig1} demonstrate that StyleFlow can effectively treat specific degradations in different degraded images as distinct styles and transfer these styles to the content image. Based on this observation, we utilize two simple loss functions, namely content loss and style loss, to guide the ConStyle in avoiding the transfer of such degradations to the latent feature. As depicted in \cref{fig:fig2} (c), the latent feature is encouraged to be far away from the degradation space and as close as possible to the clean space under the influence of content, style, and contrastive loss.

In this paper, we introduce a novel plug-and-play module for IR called ConStyle (\cref{fig:fig2} (a)), based on the guidelines mentioned above and style transfer. The ConStyle is built upon the MoCo \cite{1momentum} framework, replaces the original pretext task with a more suitable one for IR, and adopts a different encoder architecture. In addition, content loss and style loss are introduced to achieve reverse style transfer. Furthermore, thanks to the flexibility of ConStyle, we develop a general U-Net restoration network for IR. Together, ConStyle and the general restoration network (\cref{fig:fig2} (b)) form the IR framework: \textbf{IRConStyle}, with the module in the restoration network being replaceable by any specific operator. The choice of a U-Net structure for the restoration network is motivated by its ability to extract multi-scale information and its consistency with many existing IR networks \cite{50tu2022maxim,51wang2022uformer,52cai2021toward,53liu2022udc,27zhang2023kbnet,23zamir2022restormer,63li2023learning,64cui2023strip}. The latent feature extracted by ConStyle, which also contains multi-scale information, integrates well with this structure. Finally, to demonstrate the capability and compatibility of ConStyle, we replace the restoration network with Restormer \cite{23zamir2022restormer} (a transformer-based network), NAFNet \cite{22chen2022simple} (a CNN-based network), and MAXIM-1S \cite{50tu2022maxim} (an MLP-based network), called ConStyle Restormer, ConStyle NAFNet and ConStyle MAXIM-1S, and conduct extensive experiments on various IR tasks including denoising, deblurring, dehazing, and deraining. The extensive experiments show that ConStyle can integrate with different types of networks and perform reliably in different IR tasks.

Our contributions can be summarized as follows:
\begin{itemize}
\item We thoroughly analyze the problems and limitations of CL in IR and propose three guidelines that can enhance the effectiveness of contrastive learning in IR.
\item We propose a general IR framework: IRConStyle consisting of ConStyle and a general restoration network. Notably, the module within the restoration network can be easily substituted with any specific operator.
\item We innovatively apply the idea of image style transfer to CL. Which significantly improves the performance of ConStyle without introducing any parameters or increasing the inference burden.
\end{itemize}

\section{Related Work} 
\label{related_work}

\subsection{Contrastive Learning}
\label{contrastive_learning}
Contrastive learning, also known as self-supervised or unsupervised learning, is a method that enables efficient pre-training on large-scale unlabeled data. It achieves this by setting up ingenious pretext tasks. This method often performs as well as, or even surpasses, the supervised counterpart on certain tasks by fine-tuning on labeled data after pre-training \cite{5caron2021emerging,6chen2020big,11chen2020simple,42oord2018representation,60zhou2021ibot,61bao2022vlmo}. The InstDisc \cite{7wu2018unsupervised} proposes Instance Discrimination as a pretext task for the first time. In this approach, a batch of images is fed into an encoder at each iteration to obtain the corresponding latent code, which is then stored in a memory bank. The contrastive loss of each iteration is calculated between the current latent code and all historical latent codes in the memory bank; InvaSpread \cite{8ye2019unsupervised} also uses Instance Discrimination as a pretext task but does not utilize a memory bank or other data structure for storage. It solely relies on an encoder for end-to-end training, thus not fully realizing the potential of contrastive learning. To address this, MoCo \cite{1momentum} leverages the queue first-in first-out property and the strategy of exponential moving average (EMA) to construct data structures that enable effective contrastive learning. This approach has shown excellent performance in high-level tasks. However, most methods \cite{17li2022drcnet,15wu2023practical,16xia2022efficient,18moon2022feature,28chen2022unpaired,34chen2022unpaired} in IR simply adopt the contrastive loss of the CL paradigm. They do not utilize large and consistent data structures to store latent codes, making it challenging to fully leverage the potential of the CL paradigm, even with large batch sizes.

\subsection{Image Restoration}
\label{iamge_restoration}
In recent years, we have seen remarkable achievement in IR \cite{36zhang2023ingredient,37huang2023contrastive,39sun2022shufflemixer,40li2020lapar,45hui2018fast,46hui2019lightweight,47liu2020residual,57ignatov2020replacing,58chen2023human}. IPT \cite{54chen2021pre} first adopts the transformer into low-level tasks to achieve image super-resolution, denoising, and deraining. At the same time, contrastive loss is adopted in IPT where patches within the same image are considered positive samples, while patches from different images are treated as negative samples. To reduce the computational complexity of the transformer, the sliding window mechanism \cite{10liu2021swin,38liang2021swinir} is introduced based on ViT \cite{9dosovitskiy2020image}; To further reduce the computation, \cite{23zamir2022restormer} propose a transformer block specifically designed for IR; Additionally, \cite{26li2023efficient,59chen2023activating} utilize a mixed attention mechanism within the transformer block to extract more informative features; However, \cite{22chen2022simple} argues that complex nonlinear operators and attention mechanisms may be unnecessary, as satisfactory results could be achieved using a simple CNN network. For \cite{21zamir2021multi}, it proposes the Multi-Stage Progressive Restoration method, which employs a U-Net architecture for the first two stages. Inspired by the remarkable success of MLP \cite{55tolstikhin2021mlp,56liu2021pay}, \cite{50tu2022maxim} leverage gMLP \cite{56liu2021pay} to extract both local and global information from the feature map. Different from them, we introduce a simple yet effective Image Restoration framework, in which the modules are replaceable by any specific operator.

\subsection{Style Transfer}
\label{style_transfer}
Image style transfer \cite{19fan2022styleflow,44gatys2016image,48yoo2019photorealistic,49xu2021drb} is a process that involves taking two input images, a content image and a style image, and extracting their respective content and style to generate a new image. This new image retains the content of the original content image while incorporating the style of the style image. CNN is adopted for the first time \cite{44gatys2016image} to realize the style transfer, and then \cite{48yoo2019photorealistic} implements style transfer through Wavelet Transforms. Since image style transfer is a generative task,  Generative Adversarial Network is applied to realize the style transfer in an unsupervised manner \cite{49xu2021drb}. In our training process, we leverage the style transfer to guide the encoder's latent feature away from degradation space and towards clean space.
\section{Method} 
\label{method}
In this section, we first provide a detailed analysis of why the CL paradigm is not effective in IR and draw corresponding conclusions. Subsequently, we introduce the entire IRConStyle. Finally, we demonstrate how ConStyle integrates with Restormer, NAFNet, and MAXIM-1S.

\subsection{Analysis of Contrastive Learning in IR}
\label{Analysis}
\textbf{Limitations of CL paradigm in IR.} We believe that the main reason why CL is not as widely used in low-level tasks compared to high-level tasks is due to the difference in the output of the network. In high-level tasks, the network outputs an abstract semantic feature, while in low-level tasks, the output is an image of the same size as the input. Specifically, the network in high-level tasks does not have strict requirements on the size of the input image and can output an n-dimensional feature vector. However, in low-level tasks, the input and output of the network usually need to remain the same size or even require the output image to be several times larger than the input (e.g., image super-resolution). If a single network is used for the CL paradigm, it cannot guarantee that an image of the same size as the input can be restored from the corresponding latent code. This is because the network does not know the size of the real image in advance, and the real image size can even reach 4K, leading to the dimensions of the latent code being much lower than the dimensions of the real image. Therefore, most works in low-level tasks only use contrastive losses of the CL paradigm. Another solution is to employ two networks \cite{12wang2021unsupervised,13li2022all}, one for CL and one for restoration. The former is used to extract latent code and feed it to the latter, which utilizes such latent code to help rebuild the degraded image into a clean one. It is important to note that whether these works adopt the CL paradigm or not, the final effect is not significantly improved. Here, we propose three guidelines to fully leverage the CL paradigm in IR, and ablation studies are made in \cref{gudelines}.

\begin{figure}[tbp]
\centering
\includegraphics[width=\linewidth]{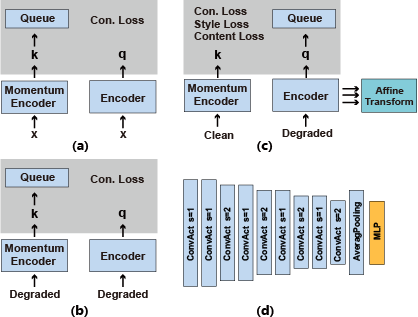}
\caption{The comparison of CL part in different networks. (a) MoCo. (b) MoCo adopted in DASR and AirNet. (c) ConStyle. (d) The momentum encoder and encoder architecture in ConStyle. Where Con. stands for Contrastive.}     
\label{fig:fig3}
\end{figure}

\textbf{Guideline 1. Additional data structures should be used to store enough positive/negative samples.} For IR, the size of the feature map in the model usually remains unchanged from input to output. This poses a challenge when using large batch sizes, such as 4096 as seen in SimCLR \cite{11chen2020simple} and BYOL \cite{3bootstrap}. The small batch size results in very few positive samples, negative samples, and queries participating in the contrastive loss calculation during each iteration. As a result, the network struggles to distinguish between positive samples, negative samples, and queries, leading to poor performance. For instance, RLFN \cite{41kong2022residual}, the winner of NTIRE 2022 \cite{43khan2022ntire}, used a batch size and patch size of 64 in its training only with contrastive loss. In its ablation experiment, the PSNR only increased by 0.01 dB on Set5, Set14, and BSD100 compared to the baseline without contrastive loss. The same situation happens in \cite{16xia2022efficient}, where the improvement over baseline is only 0.04 dB. \textbf{Conclusion: For IR, to give full play to the role of contrastive learning and avoid the huge memory occupation brought by the large batch size, additional data structures should be used to store enough positive/negative samples.} Following the MoCo, DASR and AirNet, we use the queue in our ConStyle (\cref{fig:fig3} (c)) to store the negative sample.

\textbf{Guideline 2. It is necessary to make full use of the encoder's latent feature.} Even though DASR \cite{12wang2021unsupervised} and AirNet \cite{13li2022all} adopt the complete CL paradigm (\cref{fig:fig3} (b)), the final result is not very satisfactory. This is because they do not fully utilize the latent feature of the encoder (i.e. the latent code and intermediate feature map). As mentioned earlier, both DASR and AirNet consist of a CL network and a restoration network. However, in the CL part, they only consider the highly characterized latent code as input to the restoration network. It ignores the non-highly characterized feature map, which contains hierarchical semantic information that can assist in both the restoration process and the understanding of the highly characterized latent code in the restoration network. Besides, one needs to be careful if the latent code is placed in an inappropriate position within the restoration network, as it becomes difficult to integrate it with the local feature. \textbf{Conclusion: We should not only use the latent feature but also avoid the problem of feature semantic mismatch.} Therefore, in our ConStyle, we not only utilize latent features but also incorporate them into different levels of the U-Net-style restoration network through the Affine Transform.

\textbf{Guideline 3. A reasonable pretext task should be adopted.} Another reason why DASR and AirNet perform poorly is that they use a suboptimal pretext task. In most CL paradigms, Instance Discrimination is used as a pretext task. Although this method is effective for downstream tasks like classification, segmentation, and detection, which require the network to recognize and extract semantic information. We believe it is not suitable for low-level tasks such as IR.  In IR, what we need is the ability to remove degradation and reconstruct degraded images into clean ones, rather than the ability to distinguish and extract different semantic information from images. Therefore, Instance Discrimination should not be used as DASR and AirNet did (\cref{fig:fig3} (b)). \textbf{Conclusion: Reasonable pretext tasks should be adopted in the IR domain.} In ConStyle, we input clean images to the momentum encoder and set the output as positive samples, while degraded images are input to the encoder and set the output as queries. Then, the obtained queries are fed into the queue, which is then set as the negative samples.

\begin{figure*}
\centering
\includegraphics[width=\linewidth]{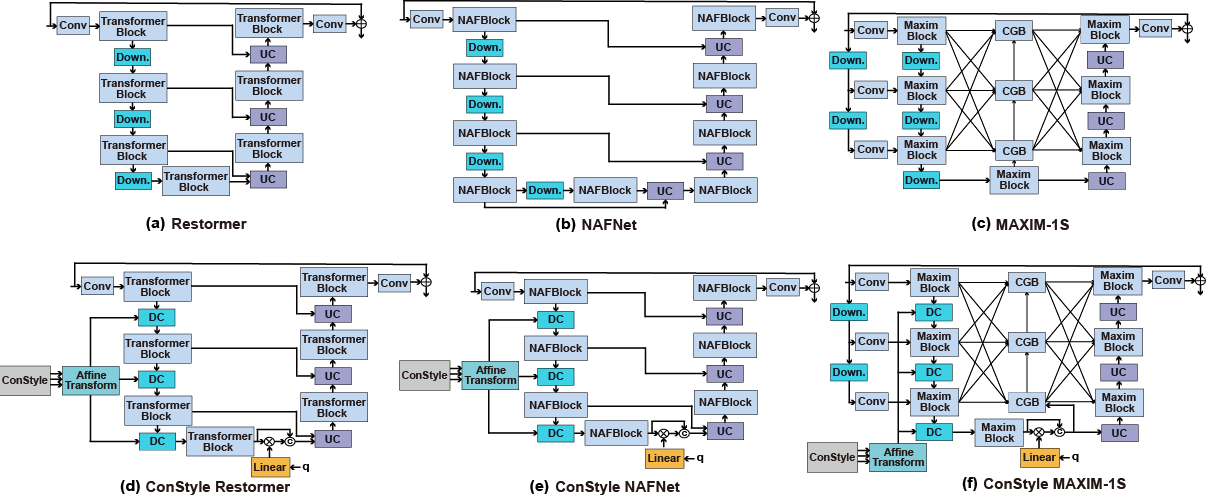}
\caption{The detailed structure of the original models (a)(b)(c) and the ConStyle models (d)(e)(f). DC represents the downsample and concat operation, and UC represents upsample and concat operation.}
\label{fig:fig4}
\end{figure*}

\subsection{Image Restoration Framework: IRConStyle}
\label{problem_formulation}
IRConStyle (\cref{fig:fig2}) consists of two main components: ConStyle and the general restoration network. ConStyle is responsible for extracting latent features and passing them to the restoration network, which is responsible for the restoration process. The training and inference flow of our IRConStyle can be described as follows:
\begin{equation}
I_{restored}=G(E(I_{degraded}, I_{clean}), I_{degraded})
\label{equ1}
\end{equation}
\begin{equation}
I_{restored}=G(E(I_{degraded}), I_{degraded})
\label{equ2}
\end{equation}

Where $E$ is the ConStyle, $G$ is the general restoration network, $I_{degraded}$ is the degraded image, $I_{clean}$ is the clean image corresponding to $I_{degraded}$, and $I_{restored}$ is the restored clean image. In ConStyle, we employ both the encoder and momentum encoder in the training (as described in \cref{equ1}), and remove the momentum encoder in the inference (as described in \cref{equ2}). To ensure that the latent features are far from the degradation space and closer to the clean space, the content and style loss are introduced. The total loss in our IRConStyle is:
\begin{equation}
L_{total}=L_{style}+L_{content}+L_{InfoNCE}+L_1
\label{equ3}
\end{equation}

$L_1$ loss is a commonly used loss function in the IR to measure the distance between $I_{clean}$ and $I_{restored}$. We will describe the style loss ($L_{style}$), content loss ($L_{content}$), and InfoNCE loss ($L_{InfoNCE}$) in detail in the next section.

\subsection{ConStyle}
\label{ConStyle}
Following MoCo, DASR, and AirNet, our encoder and momentum encoder (\cref{fig:fig3} (d)) is a siamese network consisting of several convolutions and an MLP. We also update the encoder by backpropagation and use EMA to slowly update the momentum encoder through the encoder. As analyzed before, the use of Instance Discrimination is detrimental to IR. In DASR and AirNet (\cref{fig:fig3} (b)), degraded images are fed into the momentum encoder and encoder separately, with the output $k$ representing the positive sample and the output $q$ representing the query. Each iteration adds the obtained $k$ to the queue, and then the entire queue is treated as negative samples. In contrast, in our ConStyle, we input clean images to the momentum encoder and set the output $k$ as the positive sample, while degraded images are fed into the encoder, and the output $q$ is set as the query. Each iteration adds the obtained $q$ to the queue, and then the entire queue is considered negative sample. Consistent with previous work \cite{1momentum,2improved,12wang2021unsupervised,13li2022all}, we also use InfoNCE as our contrastive loss:
\begin{equation}
L^{DASR, AirNet}_{InfoNCE}=-\log\frac{\exp({q{\cdot}k^\prime/t})}{\sum_{i=0}^N\exp({q{\cdot}Q_{i}/t})}
\label{equ4}
\end{equation}
\begin{equation}
L^{ConStyle}_{InfoNCE}=-\log\frac{\exp({q{\cdot}k/t})}{\sum_{i=0}^N\exp({q{\cdot}Q_{i}/t})}
\label{equ5}
\end{equation}

\cref{equ4} represents the $L_{InfoNCE}$ used in DASR and AirNet, while \cref{equ5} represents the $L_{InfoNCE}$ used in ConStyle. In these equations, $t$ represents a fixed temperature hyper-parameter, $Q$ represents the queue, and $N$ represents the queue size. $k^{\prime}$ is obtained by inputting $I_{degraded}$ into the momentum encoder, $k$ is obtained by inputting $I_{clean}$ into the momentum encoder, and $q$ is obtained by inputting $I_{degraded}$ into the encoder. $\sum_{i=0}^N\exp({q{\cdot}Q_{i}/t})$ indicates the computation between query and negative samples in the queue at each iteration.

\begin{table*}[htbp]
\renewcommand\arraystretch{0.8}
\centering
\setlength\tabcolsep{11pt} 
\scalebox{0.7}{
\begin{tabular}{ccccccccccc}
\toprule[1.3pt] 
     Method    &       & Down. 1 & Down. 2 & Down. 3 & Down. 4 & Up. 4 & Up. 3  & Up. 2  & Up. 1  & Total   \\ \midrule[0.8pt]
\multicolumn{1}{c|}{\multirow{2}{*}{Restormer \cite{23zamir2022restormer} }}          & Params(K) & 10.3    & 41.4    & 165.8   & 0       & 0     & 2654.2 & 663.5  & 165.8  & 3701.3  \\
\multicolumn{1}{c|}{}                                    & GFLOPs    & 0.339   & 0.339   & 0.339   & 0       & 0     & 1.358  & 1.358  & 1.358  & 5.091   \\ \midrule[0.8pt]
\multicolumn{1}{c|}{\multirow{2}{*}{ConStyle Restormer}} & Params(K) & 5.1     & 20.7    & 46.6    & 0       & 0     & 801.7  & 1368.5 & 764.9  & 3007.8  \\
\multicolumn{1}{c|}{}                                    & GFLOPs    & 0.169   & 0.169   & 0.095   & 0       & 0     & 0.495  & 3.056  & 6.715  & 10.699  \\ \midrule[0.8pt]
\multicolumn{1}{c|}{\multirow{2}{*}{NAFNet \cite{22chen2022simple}}}             & Params(K) & 32.8    & 131.3   & 524.8   & 2098.1  & 32.7  & 131.0  & 524.2  & 2097.1 & 5572.4 \\
\multicolumn{1}{c|}{}                                    & GFLOPs    & 0.268   & 0.268   & 0.268   & 0.268   & 0.267 & 0.267  & 0.268  & 0.268  & 2.142   \\ \midrule[0.8pt]
\multicolumn{1}{c|}{\multirow{2}{*}{ConStyle NAFNet}}    & Params(K) & 5.1     & 20.7    & 46.6    & 0       & 0     & 801.7  & 1368.5 & 764.9  & 3007.8 \\
\multicolumn{1}{c|}{}                                    & GFLOPs    & 0.169   & 0.169   & 0.095   & 0       & 0     & 0.495  & 3.056  & 6.715  & 10.699  \\ \midrule[0.8pt]
\multicolumn{1}{c|}{\multirow{2}{*}{MAXIM-1S \cite{50tu2022maxim}}}              & Params(K) & 16.4    & 65.6    & 262.2   & 0       & 0     & 65.6   & 32.8   & 8.2    & 450.9   \\
\multicolumn{1}{c|}{}                                    & GFLOPs    & 0.134   & 0.134   & 0.134   & 0       & 0     & 0.033  & 0.067  & 0.067  & 0.569   \\ \midrule[0.8pt]
\multicolumn{1}{c|}{\multirow{2}{*}{ConStyle MAXIM-1S}}     & Params(K) & 2.3     & 9.2     & 20.7    & 0       & 0     & 331.7  & 331.7  & 147.4  & 843.2   \\
\multicolumn{1}{c|}{}                                    & GFLOPs    & 0.075   & 0.075   & 0.042   & 0       & 0     & 0.169  & 0.679  & 1.206  & 2.246   \\ \bottomrule[1.3pt]
\end{tabular}
}
\caption{Comparison of the parameters and computations generated by using different upsample (Up.) and downsample (Down.) operation for ConStyle models and original models.}
\label{upsample_downsample}
\end{table*}

\begin{table}[h]
\renewcommand\arraystretch{0.8}
\small
\centering
\scalebox{0.95}{
\begin{tabular}{cccc}
\toprule[1.3pt]  Method                                  & \multicolumn{1}{l}{Params (M)} & \multicolumn{1}{l}{GFLOPs} & \multicolumn{1}{l}{Speed (us)} \\ \midrule[0.8pt]
\multicolumn{1}{c}{Restormer \cite{23zamir2022restormer}}          & 26.12                          & 70.49                      & 60                            \\
\multicolumn{1}{c}{ConStyle Restormer} & 15.57                          & 74.92                      & 61                             \\ \midrule[0.8pt]
\multicolumn{1}{c}{NAFNet \cite{22chen2022simple}}             & 87.40                          & 49.13                      & 53                             \\
\multicolumn{1}{c}{ConStyle NAFNet}    & 12.74                          & 46.97                      & 54                             \\ \midrule[0.8pt]
\multicolumn{1}{c}{MAXIM-1S \cite{50tu2022maxim}}              & 8.17                           & 21.58                      & 53                             \\  
\multicolumn{1}{c}{ConStyle MAXIM-1S}     & 8.10                           & 25.58                      & 52                             \\ \bottomrule[1.3pt]
\end{tabular}}
\caption{Model (w/wo ConStyle) parameters, computation, and inference speed comparison.}
\label{parametes}
\end{table}

Following guideline 3, we not only input the latent code $q$ into the restoration network but also input the feature map generated in the encoder into it using the Affine Transform. Because the feature map in the encoder is not as highly characterized as the latent code q. If only highly characterized $q$ is fed into the restoration network, it is difficult for the latter to fuse $q$ with local features, causing the problem of feature semantic mismatch. Furthermore, we leverage the idea of style transfer to guide the encoder in producing latent features that are closer to the clean space and further away from the degradation space. Specifically, we set up content loss and style loss to ensure this process:
\begin{equation}
L_{style}=-(\|q^{T}_{1}q_{1}-q^{T}q\|_2+\|q^{T}_{2}q_{2}-q^{T}q\|_2)
\end{equation}
\begin{equation}
L_{content}=\|k^{T}k-q^{T}q\|_2
\end{equation}

In each iteration, there are both incoming and outgoing latent codes in the queue. $q_{1}$ represents the latent code that is currently coming out of the queue, and $q_{2}$ represents the latent code that is about to come out of the queue. In \cite{19fan2022styleflow,44gatys2016image}, VGG \cite{62simonyan2014very} and gram matrix are used to extract style and content features. Then MSE is used to calculate style loss and content loss. Instead, our ConStyle, with the assistance of highly characterized $k$ and $q$, directly computes the corresponding content loss and style loss through MSE. This approach brings significant benefits at a minor training cost and without any inference burden.

\subsection{ConStyle with Other Network}
\label{other}
To demonstrate the versatility of our IRConStyle, we replace the restoration network with other state-of-the-art networks: Restormer \cite{23zamir2022restormer} (transformer-based), NAFNet \cite{22chen2022simple} (CNN-based), and MAXIM-1S \cite{50tu2022maxim} (MLP-based) (\cref{fig:fig4}). Restormer, NAFNet, and MAXIM-1S use different operators to implement the downsample and upsample modules. For instance, Restormer utilizes PixelUnShuffle for downsample and PixelShuffle for upsample, whereas MAXIM-1S uses stride Conv for downsample and ConvTranspose for upsample. In ConStyle Restormer, ConStyle NAFNet, and ConStyle MAXIM-1S, we use unified downsample and upsample modules of the proposed general restoration network, where the downsample consists of 1$\times$1 Conv and PixelUnShuffle, while the upsample consists of 1$\times$1 Conv and PixelShuffle.

To highlight the powerful capabilities of ConStyle and demonstrate that the improvement of the ConStyle models is not brought by simply expanding the scale of the network, we downscale ConStyle models by reducing the width or depth. Because of the introduction of ConStyle part, the parameters of networks will be increased by 1.19M. For instance, in NAFNet, the number of blocks in U-Net left, U-Net bottom, and U-Net right is [5, 6, 7, 7], 7, and [7, 7, 6, 5] respectively, with a network width of 64. In ConStyle NAFNet, the number of blocks in U-Net left, U-Net bottom, and U-Net right is [7, 8, 9], 9, and [9, 8, 7] respectively, with a network width of 48. Despite the additional 1.19M parameters introduced by ConStyle, our ConStyle NAFNet has 74.66M (85\%) fewer than NAFNet.

\section{Experiments} 
\label{experiments}

\begin{table*}[htbp]
\renewcommand\arraystretch{0.9}
\centering
\scalebox{0.75}{
\begin{tabular}{ccccccccccccc}
\toprule[1.3pt]
\multirow{2}{*}{Method} & \multicolumn{3}{c}{CBSD68($*$) \cite{bsd68}} & \multicolumn{3}{c}{Urban 100($*$) \cite{urban100}} & \multicolumn{3}{c}{CBSD68 \cite{bsd68}} & \multicolumn{3}{c}{Urban 100 \cite{urban100}} \\ 
\cmidrule{2-13} & $\sigma$=15 & $\sigma$=25 & $\sigma$=50 & $\sigma$=15 & $\sigma$=25 & $\sigma$=50 & $\sigma$=15 & $\sigma$=25 & $\sigma$=50 & $\sigma$=15 & $\sigma$=25 & $\sigma$=50 \\
\midrule[0.9pt]
Restormer \cite{23zamir2022restormer}                & 34.32                   & 31.71               & 28.51  & 34.87  & 32.65  & 29.63  & 34.37  & 31.74  & 28.53  & 35.00  & 32.75  & 29.71  \\
ConStyle Restormer       & 34.33                   & 31.71               & 28.51  & 34.89  & 32.66  & 29.64  & 34.37  & 31.74  & 28.52  & 35.01  & 32.74  & 29.71  \\
Differ.                  & {\color[HTML]{3166FF} +0.01} & {\color[HTML]{3166FF} 0} & {\color[HTML]{3166FF} 0}     & {\color[HTML]{3166FF} +0.02} & {\color[HTML]{3166FF} +0.01} & {\color[HTML]{3166FF} +0.01} & {\color[HTML]{3166FF} 0}           & {\color[HTML]{3166FF} +0} & {\color[HTML]{3166FF} +0} & {\color[HTML]{FE0000} -0.01}       & {\color[HTML]{FE0000} -0.01} & {\color[HTML]{3166FF} +0} \\ 
\midrule[0.9pt]
NAFNet \cite{22chen2022simple}                   & 34.26                   & 31.64               & 28.46  & 34.62  & 32.36  & 29.29  & 34.32  & 31.71  & 28.52  & 34.80  & 32.65  & 29.63  \\
ConStyle NAFNet          & 34.31                   & 31.69               & 28.50  & 34.82  & 32.58  & 29.55  & 34.34  & 31.71  & 28.52  & 34.91  & 32.64  & 29.63  \\
Differ.                  & {\color[HTML]{3166FF} +0.05}                  & {\color[HTML]{3166FF} +0.05}             & {\color[HTML]{3166FF} +0.04} & {\color[HTML]{3166FF} +0.20} & {\color[HTML]{3166FF} +0.22} & {\color[HTML]{3166FF} +0.26}  & {\color[HTML]{3166FF} +0.02} & {\color[HTML]{3166FF} 0}     & {\color[HTML]{3166FF} +0} & {\color[HTML]{3166FF} +0.11} & {\color[HTML]{FE0000} -0.01} & {\color[HTML]{3166FF} +0} \\ 
\midrule[0.9pt]
MAXIM-1S \cite{50tu2022maxim}                    & 34.25                   & 31.63               & 28.43  & 34.52  & 32.23  & 29.10  & 34.28  & 31.65  & 28.44  & 34.63  & 32.31  & 29.14  \\
ConStyle MAXIM-1S           & 34.25                   & 31.63               & 28.43  & 34.56  & 32.26  & 29.10  & 34.28  & 31.65  & 28.44  & 34.64  & 32.32  & 29.13  \\
Differ.                  & {\color[HTML]{3166FF} 0}                     & {\color[HTML]{3166FF} 0}                  & {\color[HTML]{3166FF} 0}     & {\color[HTML]{3166FF} +0.04} & {\color[HTML]{3166FF} +0.03} & {\color[HTML]{3166FF} 0}     & {\color[HTML]{3166FF} 0}           & {\color[HTML]{3166FF} 0}           & {\color[HTML]{3166FF} 0}     & {\color[HTML]{3166FF} +0.01} & {\color[HTML]{3166FF} +0.01} & {\color[HTML]{FE0000} -0.01} \\ 
\bottomrule[1.3pt]
\end{tabular}}
\caption{RGB image denoising tested by PSNR. Where ($*$) indicates models are trained with random sigma (0 to 50), while other indicates models are trained with fixed sigma. Blue means better and red means worse.}
\label{denoising}
\end{table*}

\begin{table*}[htbp]
\renewcommand\arraystretch{0.9}
\centering
\scalebox{0.8}{
\begin{tabular}{ccccc}
\toprule[1.3pt]
Method             & GoPro \cite{gopro}         & RealBlur-R \cite{realblur}   & RealBlur-J \cite{realblur}   & HIDE \cite{hide}          \\ \midrule[0.8pt]
Restormer \cite{23zamir2022restormer}          & 31.28/0.9188  & 33.92/0.9446  & 26.61/0.8286  & 30.16/0.9097  \\
ConStyle Restormer & 31.45/0.9208  & 33.94/0.9454  & 26.63/0.8288  & 30.02/0.9098  \\
Differ.            & {\color[HTML]{3166FF} +0.17}/{\color[HTML]{3166FF} +0.0020} & {\color[HTML]{3166FF} +0.02}/{\color[HTML]{3166FF} +0.0008} & {\color[HTML]{3166FF} +0.02}/{\color[HTML]{3166FF} +0.0002} & {\color[HTML]{3166FF} +0.04}/{\color[HTML]{3166FF} +0.0001} \\ \midrule[0.8pt]
NAFNet \cite{22chen2022simple}             & 31.29/0.9193  & 33.75/0.9434  & 26.62/0.8219  & 30.11/0.9095  \\
ConStyle NAFNet    & 31.56/0.9230  & 33.89/0.9441  & 26.61/0.8308  & 30.24/0.9106  \\
Differ.            & {\color[HTML]{3166FF} +0.27}/{\color[HTML]{3166FF} +0.0037} & {\color[HTML]{3166FF} +0.14}/{\color[HTML]{3166FF} +0.0007} & {\color[HTML]{FE0000} -0.01}/{\color[HTML]{3166FF} +0.0089} & {\color[HTML]{3166FF} +0.13}/{\color[HTML]{3166FF} +0.0011} \\ \midrule[0.8pt]
MAXIM-1S \cite{50tu2022maxim}              & 30.50/0.9039  & 33.89/0.9431  & 26.48/0.8230  & 29.22/0.8918  \\
ConStyle MAXIM-1S     & 30.77/0.9093  & 33.93/0.9435  & 26.54/0.8261  & 29.26/0.8951  \\
Differ.            & {\color[HTML]{3166FF} +0.27}/{\color[HTML]{3166FF} +0.0054} & {\color[HTML]{3166FF} +0.04}/{\color[HTML]{3166FF} +0.0004} & {\color[HTML]{3166FF} +0.06}/{\color[HTML]{3166FF} +0.0031} & {\color[HTML]{3166FF} +0.04}/{\color[HTML]{3166FF} +0.0033}    \\ \bottomrule[1.3pt]
\end{tabular}
}
\caption{Image deblurring tested by PSNR/SSIM. Blue means better and red means worse.}
\label{deblurring}
\end{table*}

\subsection{Experimental Settings}
\label{experimental_settings}
All experiments in this paper are performed on an NVIDIA Tesla V100 GPU. We set the patch size as 128 in all training. Due to memory size limitations, ConStyle NAFNet, ConStyle MAXIM-1S, NAFNet, and MAXIM-1S use a batch size of 16, while ConStyle Restormer and Restormer use a batch size of 8.  We use the AdamW ($\beta_{1}$=0.9, $\beta_{2}$=0.999) optimizer with an initial learning rate of $3e^{-4}$ and employ only random cropping and flipping as data augmentation strategies. Cosine annealing (from $3e^{-4}$ to $1e^{-6}$) and weight decay ($1e^{-4}$) are also used. ConStyle models are trained and original models are retrained, under the same training setting. We decide to retrain Restormer, NAFNet, and MAXIM-1S for two reasons. On the one hand, eliminate the differences caused by different training strategies, such as Restormer's progressive learning approach and NAFNet’s big batch size; On the other hand, we lack enough computing resources to support ConStyle models using the original models' training setting. In Restormer, for example, patch size and batch size are set to (160$^2$,40), forcing us to need at least six V100 GPUs for training. The same situation is true for NAFNet.

Following the Restormer, NAFNet, and MAXIM-1S, for deblurring, we train all models using GoPro \cite{gopro} and evaluate them on the GoPro, RealBlur-R \cite{realblur}, RealBlur-J \cite{realblur}, and HIDE \cite{hide}; For denoising, DFWB (DIV2K \cite{div2k}, Flickr2K, WED \cite{wed}, and BSD500 \cite{bsd500}) are used for training, and CBSD68 \cite{bsd68} and Urban100 \cite{urban100} are used for evaluation; For dehazing, OTS \cite{ots_sots} is used for training and SOTS \cite{ots_sots} outdoor is for evaluation; For deraining, Rain14000 \cite{fu2017removing} and Rain100H \cite{rain100l} are used for training and Rain100H and Rain100L \cite{rain100l} are for evaluation.

\subsection{Model Analyses}
\label{Model_analysis}
\cref{parametes} presents a comparison of parameters, computation, and speed between the original models and our ConStyle models. All the results are obtained using input data of size (2,3,128,128), and the speed is the average of 10,000 inference. As shown in \cref{parametes}, despite the introduction of additional parameters (1.19M), ConStyle models have fewer parameters than the original models. While the GFLOPs may not accurately measure the inference burden, we provide the inference time. The decrease in the number of parameters and increase in computation may seem counterintuitive. For instance, ConStyle Restormer reduces the number of parameters by 10.55 M but increases the computation by about 3 GFLOPs. This discrepancy arises from the use of different operators for downsampling and upsampling. As discussed in \cref{other}, in our ConStyle models, we replace the original models' downsampling and upsampling with those from the restoration network. \cref{upsample_downsample} demonstrates that this replacement adds 5.64, 8.56, and 1.67 GFLOPs to ConStyle Restormer, ConStyle NAFNet, and ConStyle MAXIM-1S, respectively.

\begin{figure*}[htbp]
\includegraphics[width=\linewidth]{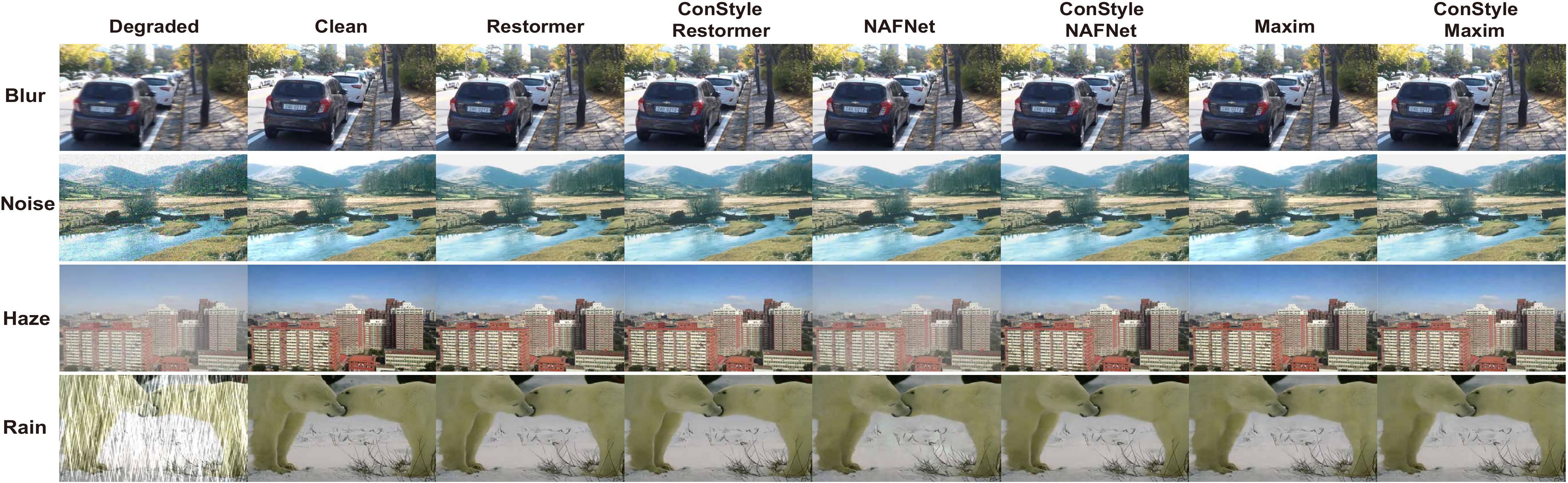}
\caption{The comparison of deblurring, dehazing, denoising, and deraining between ConStyle models and original models.}   
\label{fig:origin}
\end{figure*}

\subsection{Experiment Results}
\label{Experiment_Results}

\textbf{Image Deblurring.} ConStyle NAFNet, ConStyle MAXIM-1S, NAFNet, and MAXIM-1S are trained for 100 epochs, while ConStyle Restormer and Restormer are for 75 epochs. As shown in \cref{deblurring}, except for the slightly worse performance of ConStyle NAFNet on the RealBlur-J and ConStyle MAXIM-1S on the RealBlur-R, in general, our ConStyle models can improve the performance of original models with the help of ConStyle. For example, ConStyle NAFNet significantly outperforms NAFNet on GoPro, RealBlur-R, and HIDE, with improvements of 0.27 dB, 0.14 dB, and 0.13 dB on PSNR, respectively. ConStyle Restormer also significantly outperforms Restormer on the GoPro, with a PSNR increase of 0.17 dB.

\textbf{Image Denosing.} ConStyle NAFNet, ConStyle MAXIM-1S, NAFNet, and MAXIM-1S are trained for 100 epochs, while ConStyle Restormer and Restormer are for 75 epochs. \cref{denoising} demonstrates that in the random sigma benchmarks, the PSNR of all ConStyle models is higher than or equal to the corresponding original models. It's worth noting that, with 85\% fewer parameters, ConStyle NAFNet significantly outperforms NAFNet on the Urban100 (+0.20, +0.22, and +0.26 dB).

\textbf{Image Dehazing.} ConStyle NAFNet, ConStyle MAXIM-1S, NAFNet, and MAXIM-1S are trained for 21 epochs, while ConStyle Restormer and Restormer are trained for 16 epochs. \cref{dehazing_deraining} demonstrates that ConStyle Restormer performs slightly better than Restormer, while ConStyle MAXIM-1S performs slightly worse than MAXIM-1S. Notably, ConStyle NAFNet achieves a significant performance boost of 4.16 dB compared to NAFNet.

\textbf{Image Deraining.} ConStyle NAFNet, ConStyle MAXIM-1S, NAFNet, and MAXIM-1S are trained for 46 epochs, while ConStyle Restormer and Restormer are trained for 35 epochs. \cref{dehazing_deraining} demonstrates that except that ConStyle MAXIM-1S performs worse than MAXIM-1S on the Rain100H, all ConStyle models significantly outperform the original models on all benchmarks.

\begin{table}[tbp]
\centering
\scalebox{0.7}{
\begin{tabular}{cccc}
\toprule[1.1pt]
Method             & SOTS outdoor \cite{ots_sots}          & Rain100H \cite{rain100l}      & Rain100L \cite{rain100l}      \\ \midrule[0.8pt]
Restormer \cite{23zamir2022restormer}          & 30.83/0.9761  & 30.46/0.9047  & 37.82/0.9832  \\
ConStyle Restormer & 30.85/0.9760  & 30.82/0.9095  & 38.23/0.9844  \\
Differ.            & {\color[HTML]{3166FF} +0.02}/{\color[HTML]{FE0000} -0.0001} & {\color[HTML]{3166FF} +0.36}/{\color[HTML]{3166FF} +0.0038} & {\color[HTML]{3166FF} +0.41}/{\color[HTML]{3166FF} +0.0012} \\ \midrule[0.8pt]
NAFNet \cite{22chen2022simple}             & 26.57/0.9513  & 26.94/0.8455  & 34.97/0.9685  \\
ConStyle NAFNet    & 30.73/0.9743  & 30.52/0.9061  & 37.79/0.9832  \\
Differ.            & {\color[HTML]{3166FF} +4.16}/{\color[HTML]{3166FF} +0.0230} & {\color[HTML]{3166FF} +3.58}/{\color[HTML]{3166FF} +0.0606} & {\color[HTML]{3166FF} +2.82}/{\color[HTML]{3166FF} +0.0147} \\ \midrule[0.8pt]
MAXIM-1S \cite{50tu2022maxim}              & 30.47/0.9736  & 29.50/0.8872  & 35.54/0.9717  \\
ConStyle MAXIM-1S     & 30.59/0.9713  & 29.20/0.8845  & 36.23/0.9748  \\
Differ.            & {\color[HTML]{3166FF} +0.12}/{\color[HTML]{FE0000} -0.0023} & {\color[HTML]{FE0000} -0.30}/{\color[HTML]{FE0000} -0.0027} & {\color[HTML]{3166FF} +0.69}/{\color[HTML]{3166FF} +0.0031} \\ \bottomrule[1.1pt]
\end{tabular}
}
\caption{Image dehazing and deraining tested by PSNR/SSIM. Blue means better and red means worse. The first column represents dehazing, and the others represent deraining.}
\label{dehazing_deraining}
\end{table}

\begin{table}[ht]
\centering
\scalebox{0.7}{
\begin{tabular}{cccccccc}
\toprule[1.1pt]
Content              & Style                & InfoNCE              & PSNR  & Content              & Style                & InfoNCE              & PSNR                      \\ \midrule[0.7pt]
                          &                           &                           & 31.45 &                           &                           &                           & \multicolumn{1}{l}{32.36} \\
 \Checkmark &  \Checkmark &                           & 29.98 &  \Checkmark &  \Checkmark &                           & \multicolumn{1}{l}{31.61} \\
                          &                           &  \Checkmark & 31.50 &                           &                           &  \Checkmark & \multicolumn{1}{l}{29.70} \\
 \Checkmark &  \Checkmark &  \Checkmark & 31.56 &  \Checkmark &  \Checkmark &  \Checkmark & \multicolumn{1}{l}{32.58} \\ \midrule[0.7pt]
                          &                           &                           & 26.94 &                           &                           &                           & \multicolumn{1}{l}{26.57} \\
 \Checkmark &  \Checkmark &                           & 27.60 &  \Checkmark &  \Checkmark &                           & \multicolumn{1}{l}{28.29} \\
                          &                           &  \Checkmark & 27.51 &                           &                           &  \Checkmark & \multicolumn{1}{l}{29.87} \\
 \Checkmark &  \Checkmark &  \Checkmark & 30.52 &  \Checkmark &  \Checkmark &  \Checkmark & \multicolumn{1}{l}{30.73} \\ \bottomrule[1.3pt] 
\end{tabular}}
\caption{Ablation studies on ConStyle NAFNet to evaluate the performance of loss functions. The table in the top left, top right, bottom left, and bottom right represent the results of deblurring on GoPro \cite{gopro}, denoising on Urban100 \cite{urban100} ($\sigma$=25), deraining on Rain100H \cite{rain100l}, and dehazing on SOTS outdoors \cite{ots_sots}, respectively.}
\label{ablation}
\end{table}

\begin{table}[htbp]
\centering
\scalebox{0.85}{
\begin{tabular}{ccccc}
\toprule[1.1pt]
     & Baseline & Guideline 1 & Guideline 2 & Guideline 3 \\ \midrule[0.7pt]
PSNR & 31.56    & 28.82       & 31.33       & 31.45       \\ \bottomrule[1.3pt]
\end{tabular}
}
\caption{Ablation studies of ConStyle NAFNet on GoPro \cite{gopro} to evaluate proposed three guidelines.}
\label{gudelines}
\end{table}

\subsection{Ablation Studies}
\label{ablation_studies}

We conduct two ablation experiments to evaluate the performance of loss functions (\cref{ablation}) and proposed three guidelines (\cref{gudelines}). In \cref{gudelines}, our ConStyle NAFNet is set as a baseline, in which the content, style, and InfoNCE loss are used as well as the latent feature map. For guideline 1, we change the Queue size of the ConStyle from 65760 to 16; For guideline 2, the latent feature maps are removed. For guideline3, we move the Queue from behind the Encoder back to behind the Momentum Encoder, in line with AirNet \cite{13li2022all}, DASR \cite{12wang2021unsupervised}, and MoCo \cite{1momentum}.

\section{Conclusion} 
\label{conclusion}
In this paper, we analyze the limitations of the CL paradigm in IR and propose three guidelines to fully leverage CL in IR. In addition, we introduce an image restoration framework: IRConStyle, consisting of ConStyle and the restoration network. ConStyle extracts latent features and passes them to the restoration network which is responsible for image restoration.

{\small
\bibliographystyle{ieee_fullname}
\bibliography{egbib}

\begin{thebibliography}{10}\itemsep=-1pt

\bibitem{div2k}
Eirikur Agustsson and Radu Timofte.
\newblock Ntire 2017 challenge on single image super-resolution: Dataset and
  study.
\newblock In {\em Proceedings of the IEEE conference on computer vision and
  pattern recognition workshops}, pages 126--135, 2017.

\bibitem{bsd500}
Pablo Arbelaez, Michael Maire, Charless Fowlkes, and Jitendra Malik.
\newblock Contour detection and hierarchical image segmentation.
\newblock {\em IEEE transactions on pattern analysis and machine intelligence},
  33(5):898--916, 2010.

\bibitem{61bao2022vlmo}
Hangbo Bao, Wenhui Wang, Li Dong, Qiang Liu, Owais~Khan Mohammed, Kriti
  Aggarwal, Subhojit Som, Songhao Piao, and Furu Wei.
\newblock Vlmo: Unified vision-language pre-training with
  mixture-of-modality-experts.
\newblock {\em Advances in Neural Information Processing Systems},
  35:32897--32912, 2022.

\bibitem{52cai2021toward}
Haoming Cai, Jingwen He, Yu Qiao, and Chao Dong.
\newblock Toward interactive modulation for photo-realistic image restoration.
\newblock In {\em Proceedings of the IEEE/CVF Conference on Computer Vision and
  Pattern Recognition}, pages 294--303, 2021.

\bibitem{5caron2021emerging}
Mathilde Caron, Hugo Touvron, Ishan Misra, Herv{\'e} J{\'e}gou, Julien Mairal,
  Piotr Bojanowski, and Armand Joulin.
\newblock Emerging properties in self-supervised vision transformers.
\newblock In {\em Proceedings of the IEEE/CVF international conference on
  computer vision}, pages 9650--9660, 2021.

\bibitem{58chen2023human}
Du Chen, Jie Liang, Xindong Zhang, Ming Liu, Hui Zeng, and Lei Zhang.
\newblock Human guided ground-truth generation for realistic image
  super-resolution.
\newblock In {\em Proceedings of the IEEE/CVF Conference on Computer Vision and
  Pattern Recognition}, pages 14082--14091, 2023.

\bibitem{54chen2021pre}
Hanting Chen, Yunhe Wang, Tianyu Guo, Chang Xu, Yiping Deng, Zhenhua Liu, Siwei
  Ma, Chunjing Xu, Chao Xu, and Wen Gao.
\newblock Pre-trained image processing transformer.
\newblock In {\em Proceedings of the IEEE/CVF conference on computer vision and
  pattern recognition}, pages 12299--12310, 2021.

\bibitem{22chen2022simple}
Liangyu Chen, Xiaojie Chu, Xiangyu Zhang, and Jian Sun.
\newblock Simple baselines for image restoration.
\newblock In {\em European Conference on Computer Vision}, pages 17--33.
  Springer, 2022.

\bibitem{11chen2020simple}
Ting Chen, Simon Kornblith, Mohammad Norouzi, and Geoffrey Hinton.
\newblock A simple framework for contrastive learning of visual
  representations.
\newblock In {\em International conference on machine learning}, pages
  1597--1607. PMLR, 2020.

\bibitem{6chen2020big}
Ting Chen, Simon Kornblith, Kevin Swersky, Mohammad Norouzi, and Geoffrey~E
  Hinton.
\newblock Big self-supervised models are strong semi-supervised learners.
\newblock {\em Advances in neural information processing systems},
  33:22243--22255, 2020.

\bibitem{30chen2022learning}
Wei-Ting Chen, Zhi-Kai Huang, Cheng-Che Tsai, Hao-Hsiang Yang, Jian-Jiun Ding,
  and Sy-Yen Kuo.
\newblock Learning multiple adverse weather removal via two-stage knowledge
  learning and multi-contrastive regularization: Toward a unified model.
\newblock In {\em Proceedings of the IEEE/CVF Conference on Computer Vision and
  Pattern Recognition}, pages 17653--17662, 2022.

\bibitem{2improved}
Xinlei Chen, Haoqi Fan, Ross Girshick, and Kaiming He.
\newblock Improved baselines with momentum contrastive learning.
\newblock {\em arXiv preprint arXiv:2003.04297}, 2020.

\bibitem{28chen2022unpaired}
Xiang Chen, Jinshan Pan, Kui Jiang, Yufeng Li, Yufeng Huang, Caihua Kong,
  Longgang Dai, and Zhentao Fan.
\newblock Unpaired deep image deraining using dual contrastive learning.
\newblock In {\em Proceedings of the IEEE/CVF Conference on Computer Vision and
  Pattern Recognition}, pages 2017--2026, 2022.

\bibitem{34chen2022unpaired}
Xiang Chen, Jinshan Pan, Kui Jiang, Yufeng Li, Yufeng Huang, Caihua Kong,
  Longgang Dai, and Zhentao Fan.
\newblock Unpaired deep image deraining using dual contrastive learning.
\newblock In {\em Proceedings of the IEEE/CVF Conference on Computer Vision and
  Pattern Recognition}, pages 2017--2026, 2022.

\bibitem{59chen2023activating}
Xiangyu Chen, Xintao Wang, Jiantao Zhou, Yu Qiao, and Chao Dong.
\newblock Activating more pixels in image super-resolution transformer.
\newblock In {\em Proceedings of the IEEE/CVF Conference on Computer Vision and
  Pattern Recognition}, pages 22367--22377, 2023.

\bibitem{4Chen_2021_ICCV}
Xinlei Chen, Saining Xie, and Kaiming He.
\newblock An empirical study of training self-supervised vision transformers.
\newblock In {\em Proceedings of the IEEE/CVF International Conference on
  Computer Vision (ICCV)}, pages 9640--9649, October 2021.

\bibitem{64cui2023strip}
Yuning Cui, Yi Tao, Luoxi Jing, and Alois Knoll.
\newblock Strip attention for image restoration.
\newblock In {\em International Joint Conference on Artificial Intelligence,
  IJCAI}, 2023.

\bibitem{9dosovitskiy2020image}
Alexey Dosovitskiy, Lucas Beyer, Alexander Kolesnikov, Dirk Weissenborn,
  Xiaohua Zhai, Thomas Unterthiner, Mostafa Dehghani, Matthias Minderer, Georg
  Heigold, Sylvain Gelly, et~al.
\newblock An image is worth 16x16 words: Transformers for image recognition at
  scale.
\newblock {\em arXiv preprint arXiv:2010.11929}, 2020.

\bibitem{32du2020learning}
Wenchao Du, Hu Chen, and Hongyu Yang.
\newblock Learning invariant representation for unsupervised image restoration.
\newblock In {\em Proceedings of the ieee/cvf conference on computer vision and
  pattern recognition}, pages 14483--14492, 2020.

\bibitem{19fan2022styleflow}
Weichen Fan, Jinghuan Chen, Jiabin Ma, Jun Hou, and Shuai Yi.
\newblock Styleflow for content-fixed image to image translation.
\newblock {\em arXiv preprint arXiv:2207.01909}, 2022.

\bibitem{fu2017removing}
Xueyang Fu, Jiabin Huang, Delu Zeng, Yue Huang, Xinghao Ding, and John Paisley.
\newblock Removing rain from single images via a deep detail network.
\newblock In {\em Proceedings of the IEEE conference on computer vision and
  pattern recognition}, pages 3855--3863, 2017.

\bibitem{44gatys2016image}
Leon~A Gatys, Alexander~S Ecker, and Matthias Bethge.
\newblock Image style transfer using convolutional neural networks.
\newblock In {\em Proceedings of the IEEE conference on computer vision and
  pattern recognition}, pages 2414--2423, 2016.

\bibitem{3bootstrap}
Jean-Bastien Grill, Florian Strub, Florent Altch{\'e}, Corentin Tallec, Pierre
  Richemond, Elena Buchatskaya, Carl Doersch, Bernardo Avila~Pires, Zhaohan
  Guo, Mohammad Gheshlaghi~Azar, et~al.
\newblock Bootstrap your own latent-a new approach to self-supervised learning.
\newblock {\em Advances in neural information processing systems},
  33:21271--21284, 2020.

\bibitem{1momentum}
Kaiming He, Haoqi Fan, Yuxin Wu, Saining Xie, and Ross Girshick.
\newblock Momentum contrast for unsupervised visual representation learning.
\newblock In {\em Proceedings of the IEEE/CVF conference on computer vision and
  pattern recognition}, pages 9729--9738, 2020.

\bibitem{urban100}
Jia-Bin Huang, Abhishek Singh, and Narendra Ahuja.
\newblock Single image super-resolution from transformed self-exemplars.
\newblock In {\em Proceedings of the IEEE conference on computer vision and
  pattern recognition}, pages 5197--5206, 2015.

\bibitem{37huang2023contrastive}
Shirui Huang, Keyan Wang, Huan Liu, Jun Chen, and Yunsong Li.
\newblock Contrastive semi-supervised learning for underwater image restoration
  via reliable bank.
\newblock In {\em Proceedings of the IEEE/CVF Conference on Computer Vision and
  Pattern Recognition}, pages 18145--18155, 2023.

\bibitem{46hui2019lightweight}
Zheng Hui, Xinbo Gao, Yunchu Yang, and Xiumei Wang.
\newblock Lightweight image super-resolution with information
  multi-distillation network.
\newblock In {\em Proceedings of the 27th acm international conference on
  multimedia}, pages 2024--2032, 2019.

\bibitem{45hui2018fast}
Zheng Hui, Xiumei Wang, and Xinbo Gao.
\newblock Fast and accurate single image super-resolution via information
  distillation network.
\newblock In {\em Proceedings of the IEEE conference on computer vision and
  pattern recognition}, pages 723--731, 2018.

\bibitem{57ignatov2020replacing}
Andrey Ignatov, Luc Van~Gool, and Radu Timofte.
\newblock Replacing mobile camera isp with a single deep learning model.
\newblock In {\em Proceedings of the IEEE/CVF Conference on Computer Vision and
  Pattern Recognition Workshops}, pages 536--537, 2020.

\bibitem{43khan2022ntire}
Fahad~Shahbaz Khan and Salman Khan.
\newblock Ntire 2022 challenge on efficient super-resolution: Methods and
  results.
\newblock 2022.

\bibitem{41kong2022residual}
Fangyuan Kong, Mingxi Li, Songwei Liu, Ding Liu, Jingwen He, Yang Bai, Fangmin
  Chen, and Lean Fu.
\newblock Residual local feature network for efficient super-resolution.
\newblock In {\em Proceedings of the IEEE/CVF Conference on Computer Vision and
  Pattern Recognition}, pages 766--776, 2022.

\bibitem{13li2022all}
Boyun Li, Xiao Liu, Peng Hu, Zhongqin Wu, Jiancheng Lv, and Xi Peng.
\newblock All-in-one image restoration for unknown corruption.
\newblock In {\em Proceedings of the IEEE/CVF Conference on Computer Vision and
  Pattern Recognition}, pages 17452--17462, 2022.

\bibitem{ots_sots}
Boyi Li, Wenqi Ren, Dengpan Fu, Dacheng Tao, Dan Feng, Wenjun Zeng, and
  Zhangyang Wang.
\newblock Benchmarking single-image dehazing and beyond.
\newblock {\em IEEE Transactions on Image Processing}, 28(1):492--505, 2018.

\bibitem{17li2022drcnet}
Fei Li, Lingfeng Shen, Yang Mi, and Zhenbo Li.
\newblock Drcnet: Dynamic image restoration contrastive network.
\newblock In {\em European Conference on Computer Vision}, pages 514--532.
  Springer, 2022.

\bibitem{63li2023learning}
Guanxin Li, Jingang Shi, Yuan Zong, Fei Wang, Tian Wang, and Yihong Gong.
\newblock Learning attention from attention: Efficient self-refinement
  transformer for face super-resolution.
\newblock In {\em Proceedings of the International Joint Conference on
  Artificial Intelligence}, 2023.

\bibitem{20li2020all}
Ruoteng Li, Robby~T Tan, and Loong-Fah Cheong.
\newblock All in one bad weather removal using architectural search.
\newblock In {\em Proceedings of the IEEE/CVF conference on computer vision and
  pattern recognition}, pages 3175--3185, 2020.

\bibitem{40li2020lapar}
Wenbo Li, Kun Zhou, Lu Qi, Nianjuan Jiang, Jiangbo Lu, and Jiaya Jia.
\newblock Lapar: Linearly-assembled pixel-adaptive regression network for
  single image super-resolution and beyond.
\newblock {\em Advances in Neural Information Processing Systems},
  33:20343--20355, 2020.

\bibitem{26li2023efficient}
Yawei Li, Yuchen Fan, Xiaoyu Xiang, Denis Demandolx, Rakesh Ranjan, Radu
  Timofte, and Luc Van~Gool.
\newblock Efficient and explicit modelling of image hierarchies for image
  restoration.
\newblock In {\em Proceedings of the IEEE/CVF Conference on Computer Vision and
  Pattern Recognition}, pages 18278--18289, 2023.

\bibitem{38liang2021swinir}
Jingyun Liang, Jiezhang Cao, Guolei Sun, Kai Zhang, Luc Van~Gool, and Radu
  Timofte.
\newblock Swinir: Image restoration using swin transformer.
\newblock In {\em Proceedings of the IEEE/CVF international conference on
  computer vision}, pages 1833--1844, 2021.

\bibitem{56liu2021pay}
Hanxiao Liu, Zihang Dai, David So, and Quoc~V Le.
\newblock Pay attention to mlps.
\newblock {\em Advances in Neural Information Processing Systems},
  34:9204--9215, 2021.

\bibitem{47liu2020residual}
Jie Liu, Jie Tang, and Gangshan Wu.
\newblock Residual feature distillation network for lightweight image
  super-resolution.
\newblock In {\em Computer Vision--ECCV 2020 Workshops: Glasgow, UK, August
  23--28, 2020, Proceedings, Part III 16}, pages 41--55. Springer, 2020.

\bibitem{53liu2022udc}
Xina Liu, Jinfan Hu, Xiangyu Chen, and Chao Dong.
\newblock Udc-unet: Under-display camera image restoration via u-shape dynamic
  network.
\newblock In {\em European Conference on Computer Vision}, pages 113--129.
  Springer, 2022.

\bibitem{10liu2021swin}
Ze Liu, Yutong Lin, Yue Cao, Han Hu, Yixuan Wei, Zheng Zhang, Stephen Lin, and
  Baining Guo.
\newblock Swin transformer: Hierarchical vision transformer using shifted
  windows.
\newblock In {\em Proceedings of the IEEE/CVF international conference on
  computer vision}, pages 10012--10022, 2021.

\bibitem{wed}
Kede Ma, Zhengfang Duanmu, Qingbo Wu, Zhou Wang, Hongwei Yong, Hongliang Li,
  and Lei Zhang.
\newblock Waterloo exploration database: New challenges for image quality
  assessment models.
\newblock {\em IEEE Transactions on Image Processing}, 26(2):1004--1016, 2016.

\bibitem{bsd68}
David Martin, Charless Fowlkes, Doron Tal, and Jitendra Malik.
\newblock A database of human segmented natural images and its application to
  evaluating segmentation algorithms and measuring ecological statistics.
\newblock In {\em Proceedings Eighth IEEE International Conference on Computer
  Vision. ICCV 2001}, volume~2, pages 416--423. IEEE, 2001.

\bibitem{18moon2022feature}
HyeonCheol Moon, JinWoo Jeong, and SungJei Kim.
\newblock Feature-based adaptive contrastive distillation for efficient single
  image super-resolution.
\newblock {\em arXiv preprint arXiv:2211.15951}, 2022.

\bibitem{gopro}
Seungjun Nah, Tae Hyun~Kim, and Kyoung Mu~Lee.
\newblock Deep multi-scale convolutional neural network for dynamic scene
  deblurring.
\newblock In {\em Proceedings of the IEEE conference on computer vision and
  pattern recognition}, pages 3883--3891, 2017.

\bibitem{42oord2018representation}
Aaron van~den Oord, Yazhe Li, and Oriol Vinyals.
\newblock Representation learning with contrastive predictive coding.
\newblock {\em arXiv preprint arXiv:1807.03748}, 2018.

\bibitem{25park2023all}
Dongwon Park, Byung~Hyun Lee, and Se~Young Chun.
\newblock All-in-one image restoration for unknown degradations using adaptive
  discriminative filters for specific degradations.
\newblock In {\em Proceedings of the IEEE/CVF Conference on Computer Vision and
  Pattern Recognition}, pages 5815--5824, 2023.

\bibitem{33poirier2023robust}
Yohan Poirier-Ginter and Jean-Fran{\c{c}}ois Lalonde.
\newblock Robust unsupervised stylegan image restoration.
\newblock In {\em Proceedings of the IEEE/CVF Conference on Computer Vision and
  Pattern Recognition}, pages 22292--22301, 2023.

\bibitem{realblur}
Jaesung Rim, Haeyun Lee, Jucheol Won, and Sunghyun Cho.
\newblock Real-world blur dataset for learning and benchmarking deblurring
  algorithms.
\newblock In {\em Computer Vision--ECCV 2020: 16th European Conference,
  Glasgow, UK, August 23--28, 2020, Proceedings, Part XXV 16}, pages 184--201.
  Springer, 2020.

\bibitem{hide}
Ziyi Shen, Wenguan Wang, Xiankai Lu, Jianbing Shen, Haibin Ling, Tingfa Xu, and
  Ling Shao.
\newblock Human-aware motion deblurring.
\newblock In {\em Proceedings of the IEEE/CVF International Conference on
  Computer Vision}, pages 5572--5581, 2019.

\bibitem{62simonyan2014very}
Karen Simonyan and Andrew Zisserman.
\newblock Very deep convolutional networks for large-scale image recognition.
\newblock {\em arXiv preprint arXiv:1409.1556}, 2014.

\bibitem{39sun2022shufflemixer}
Long Sun, Jinshan Pan, and Jinhui Tang.
\newblock Shufflemixer: An efficient convnet for image super-resolution.
\newblock {\em Advances in Neural Information Processing Systems},
  35:17314--17326, 2022.

\bibitem{55tolstikhin2021mlp}
Ilya~O Tolstikhin, Neil Houlsby, Alexander Kolesnikov, Lucas Beyer, Xiaohua
  Zhai, Thomas Unterthiner, Jessica Yung, Andreas Steiner, Daniel Keysers,
  Jakob Uszkoreit, et~al.
\newblock Mlp-mixer: An all-mlp architecture for vision.
\newblock {\em Advances in neural information processing systems},
  34:24261--24272, 2021.

\bibitem{50tu2022maxim}
Zhengzhong Tu, Hossein Talebi, Han Zhang, Feng Yang, Peyman Milanfar, Alan
  Bovik, and Yinxiao Li.
\newblock Maxim: Multi-axis mlp for image processing.
\newblock In {\em Proceedings of the IEEE/CVF Conference on Computer Vision and
  Pattern Recognition}, pages 5769--5780, 2022.

\bibitem{24valanarasu2022transweather}
Jeya Maria~Jose Valanarasu, Rajeev Yasarla, and Vishal~M Patel.
\newblock Transweather: Transformer-based restoration of images degraded by
  adverse weather conditions.
\newblock In {\em Proceedings of the IEEE/CVF Conference on Computer Vision and
  Pattern Recognition}, pages 2353--2363, 2022.

\bibitem{12wang2021unsupervised}
Longguang Wang, Yingqian Wang, Xiaoyu Dong, Qingyu Xu, Jungang Yang, Wei An,
  and Yulan Guo.
\newblock Unsupervised degradation representation learning for blind
  super-resolution.
\newblock In {\em Proceedings of the IEEE/CVF Conference on Computer Vision and
  Pattern Recognition}, pages 10581--10590, 2021.

\bibitem{14wang2021towards}
Yanbo Wang, Shaohui Lin, Yanyun Qu, Haiyan Wu, Zhizhong Zhang, Yuan Xie, and
  Angela Yao.
\newblock Towards compact single image super-resolution via contrastive
  self-distillation.
\newblock {\em arXiv preprint arXiv:2105.11683}, 2021.

\bibitem{51wang2022uformer}
Zhendong Wang, Xiaodong Cun, Jianmin Bao, Wengang Zhou, Jianzhuang Liu, and
  Houqiang Li.
\newblock Uformer: A general u-shaped transformer for image restoration.
\newblock In {\em Proceedings of the IEEE/CVF conference on computer vision and
  pattern recognition}, pages 17683--17693, 2022.

\bibitem{15wu2023practical}
Gang Wu, Junjun Jiang, and Xianming Liu.
\newblock A practical contrastive learning framework for single-image
  super-resolution.
\newblock {\em IEEE Transactions on Neural Networks and Learning Systems},
  2023.

\bibitem{7wu2018unsupervised}
Zhirong Wu, Yuanjun Xiong, Stella~X Yu, and Dahua Lin.
\newblock Unsupervised feature learning via non-parametric instance
  discrimination.
\newblock In {\em Proceedings of the IEEE conference on computer vision and
  pattern recognition}, pages 3733--3742, 2018.

\bibitem{16xia2022efficient}
Bin Xia, Yucheng Hang, Yapeng Tian, Wenming Yang, Qingmin Liao, and Jie Zhou.
\newblock Efficient non-local contrastive attention for image super-resolution.
\newblock In {\em Proceedings of the AAAI Conference on Artificial
  Intelligence}, volume~36, pages 2759--2767, 2022.

\bibitem{29xie2021finding}
Liangbin Xie, Xintao Wang, Chao Dong, Zhongang Qi, and Ying Shan.
\newblock Finding discriminative filters for specific degradations in blind
  super-resolution.
\newblock {\em Advances in Neural Information Processing Systems}, 34:51--61,
  2021.

\bibitem{49xu2021drb}
Wenju Xu, Chengjiang Long, Ruisheng Wang, and Guanghui Wang.
\newblock Drb-gan: A dynamic resblock generative adversarial network for
  artistic style transfer.
\newblock In {\em Proceedings of the IEEE/CVF international conference on
  computer vision}, pages 6383--6392, 2021.

\bibitem{rain100l}
Wenhan Yang, Robby~T Tan, Jiashi Feng, Jiaying Liu, Zongming Guo, and Shuicheng
  Yan.
\newblock Deep joint rain detection and removal from a single image.
\newblock In {\em Proceedings of the IEEE conference on computer vision and
  pattern recognition}, pages 1357--1366, 2017.

\bibitem{8ye2019unsupervised}
Mang Ye, Xu Zhang, Pong~C Yuen, and Shih-Fu Chang.
\newblock Unsupervised embedding learning via invariant and spreading instance
  feature.
\newblock In {\em Proceedings of the IEEE/CVF conference on computer vision and
  pattern recognition}, pages 6210--6219, 2019.

\bibitem{48yoo2019photorealistic}
Jaejun Yoo, Youngjung Uh, Sanghyuk Chun, Byeongkyu Kang, and Jung-Woo Ha.
\newblock Photorealistic style transfer via wavelet transforms.
\newblock In {\em Proceedings of the IEEE/CVF International Conference on
  Computer Vision}, pages 9036--9045, 2019.

\bibitem{23zamir2022restormer}
Syed~Waqas Zamir, Aditya Arora, Salman Khan, Munawar Hayat, Fahad~Shahbaz Khan,
  and Ming-Hsuan Yang.
\newblock Restormer: Efficient transformer for high-resolution image
  restoration.
\newblock In {\em Proceedings of the IEEE/CVF conference on computer vision and
  pattern recognition}, pages 5728--5739, 2022.

\bibitem{21zamir2021multi}
Syed~Waqas Zamir, Aditya Arora, Salman Khan, Munawar Hayat, Fahad~Shahbaz Khan,
  Ming-Hsuan Yang, and Ling Shao.
\newblock Multi-stage progressive image restoration.
\newblock In {\em Proceedings of the IEEE/CVF conference on computer vision and
  pattern recognition}, pages 14821--14831, 2021.

\bibitem{36zhang2023ingredient}
Jinghao Zhang, Jie Huang, Mingde Yao, Zizheng Yang, Hu Yu, Man Zhou, and Feng
  Zhao.
\newblock Ingredient-oriented multi-degradation learning for image restoration.
\newblock In {\em Proceedings of the IEEE/CVF Conference on Computer Vision and
  Pattern Recognition}, pages 5825--5835, 2023.

\bibitem{35zhang2022practical}
Kai Zhang, Yawei Li, Jingyun Liang, Jiezhang Cao, Yulun Zhang, Hao Tang, Radu
  Timofte, and Luc Van~Gool.
\newblock Practical blind denoising via swin-conv-unet and data synthesis.
\newblock {\em arXiv preprint arXiv:2203.13278}, 2022.

\bibitem{27zhang2023kbnet}
Yi Zhang, Dasong Li, Xiaoyu Shi, Dailan He, Kangning Song, Xiaogang Wang,
  Hongwei Qin, and Hongsheng Li.
\newblock Kbnet: Kernel basis network for image restoration.
\newblock {\em arXiv preprint arXiv:2303.02881}, 2023.

\bibitem{60zhou2021ibot}
Jinghao Zhou, Chen Wei, Huiyu Wang, Wei Shen, Cihang Xie, Alan Yuille, and Tao
  Kong.
\newblock ibot: Image bert pre-training with online tokenizer.
\newblock {\em arXiv preprint arXiv:2111.07832}, 2021.

\end{thebibliography}
}

\end{document}